\documentclass[twocolumn]{article}
\advance\topmargin-1in\advance\oddsidemargin-1in
\evensidemargin\oddsidemargin

\makeatletter
\def\@normalsize{\@setsize\normalsize{12pt}\xpt\@xpt
\abovedisplayskip 10pt plus2pt minus5pt\belowdisplayskip \abovedisplayskip
\abovedisplayshortskip \z@ plus3pt\belowdisplayshortskip 6pt plus3pt
minus3pt\let\@listi\@listI}

\def\section{\@startsection {section}{1}{\z@}{20pt plus 2pt minus 2pt}
{8pt plus 2pt minus 2pt}{\centering\normalsize\sc
\edef\@svsec{\thesection.\ }}}
\def\thesection{\Roman{section}}

\def\subsection{\@startsection {subsection}{2}{\z@}{16pt plus 2pt minus 2pt}
{6pt plus 2pt minus 2pt}{\normalsize\sl
\edef\@svsec{\thesubsection.\ }}}
\def\thesubsection{\Alph{subsection}}

\long\def\@makecaption#1#2{
\vskip10pt\begin{center} #1 #2 \end{center}\par\vskip 1pt}
\def\fnum@figure{\raggedright{\footnotesize Fig. \thefigure }.%
\footnotesize}
\def\fnum@table{\footnotesize TABLE \thetable\\\footnotesize\sc}
\def\thetable{\Roman{table}}

\makeatother

\usepackage{amsmath,amssymb,amsfonts}
\usepackage{algorithmic}
\usepackage{graphicx}
\usepackage{textcomp}
\usepackage{xcolor}

\usepackage[square,sort,comma,numbers]{natbib}

\usepackage{multirow}
\usepackage{booktabs}
\usepackage{algorithm}
\newcommand{\pName}{RADARS}
\newcommand{\DARTS}{DNAS}
\newcommand{\todo}[1]{#1}
\newcommand{\cmtColor}[1]{\textcolor[rgb]{0.3, 0.8, 0.3}{#1}}

\begin{document}
%date not printed
\date{}

%make title
\title{\Large\textbf{\pName: Memory Efficient Reinforcement Learning Aided Differentiable Neural Architecture Search}}

%for single author
% \author{}

%for two authors
\author{
Zheyu Yan\textsuperscript{*} Weiwen Jiang\textsuperscript{$\dagger$} Xiaobo Sharon Hu\textsuperscript{*} Yiyu Shi\textsuperscript{*}\\
\textsuperscript{*}University of Notre Dame \textsuperscript{$\dagger$}George Mason University\\
}
\maketitle
\thispagestyle{empty}

% {\small\textbf{Differentiable neural architecture search (\DARTS) has been known for its efficiency in automatic machine learning tasks. 
% However, \DARTS~based methods suffer from memory usage explosion when the search space expands, which may prevent them from running successfully on even advanced GPU platforms. On the other hand, reinforcement learning (RL) based methods, while being memory efficient, are extremely time-consuming. Combining the advantages of both types of methods, this paper presents \pName, a scalable RL aided \DARTS~framework that can explore large search spaces in a fast and memory-efficient manner. \pName~iteratively adopts RL to prune undesired architecture candidates and identifies a promising subspace to carry out \DARTS. Experiments using a workstation with 
% %a RTX-2080Ti GPU card (
% 12 GB GPU memory
% show that on CIFAR-10 and ImageNet datasets with large search spaces, \pName~can achieve up to 3.41\% higher accuracy and 39\% lower arithmetic intensity with 2.5X search time reduction compared with a state-of-the-art RL-based method, while the two \DARTS~baselines cannot complete due to excessive memory consumption or search time. To the best of the authors' knowledge, this is the first \DARTS~framework that can handle large search spaces with bounded memory usage.
% }}
{\small\textbf{Differentiable neural architecture search (\DARTS) is known for its capacity in the automatic generation of superior neural networks. 
However, \DARTS~based methods suffer from memory usage explosion when the search space expands, which may prevent them from running successfully on even advanced GPU platforms. On the other hand, reinforcement learning (RL) based methods, while being memory efficient, are extremely time-consuming. Combining the advantages of both types of methods, this paper presents \pName, a scalable RL aided \DARTS~framework that can explore large search spaces in a fast and memory-efficient manner. \pName~iteratively applies RL to prune undesired architecture candidates and identifies a promising subspace to carry out \DARTS. Experiments using a workstation with 12 GB GPU memory show that on CIFAR-10 and ImageNet datasets, \pName~can achieve up to 3.41\% higher accuracy with 2.5X search time reduction compared with a state-of-the-art RL-based method, while the two \DARTS~baselines cannot complete due to excessive memory usage or search time. To the best of the authors' knowledge, this is the first \DARTS~framework that can handle large search spaces with bounded memory usage.
}}

\section{Introduction}
% \todo{Pay attention to the tense}

Deep neural networks (DNNs) have achieved state-of-the-art performance in various applications such as computer vision and machine translation. To push DNNs to achieve even higher performance, Neural Architecture Search (NAS) has been proposed to automatically identify neural architectures that can outperform their hand-crafted counterparts. Existing NAS frameworks are either based on reinforcement learning (RL), or a differentiable approach.

%based algorithms. Co-exploration of neural architectures and hardware design moves a step further to search for a neural architecture-hardware design pair that can simultaneously optimize network accuracy and hardware efficiency. 
RL-based NAS frameworks (RL-NAS) sample one neural architecture at a time and use its performance as reward/punishment to guide an RL controller~\cite{zoph2016neural}. These frameworks are memory efficient and can handle a variety of optimization objectives beyond accuracy, such as power efficiency, latency, \emph{etc.}, for hardware and neural architecture co-exploration. However, the need for sampling and training a large number of architectures leads to extremely long search time, sometimes exceeding 1,000 GPU hours. 

To solve this issue, researchers propose differentiable neural architecture search (\DARTS) that trains an over-parameterized network, named SuperNet, which contains all candidate paths and reduces the search time to tens of GPU hours. 
%there are definitely more than these papers on DNAS
Yet the inclusion of different candidate paths puts great pressure on GPU memory. Existing \DARTS~based works~\cite{liu2018darts, cai2018proxylessnas, li2020edd} carefully control their search spaces and experiment hyper-parameters (\emph{e.g.}, batch size) to keep the memory usage lower than the GPU memory size to avoid out-of-memory (OOM) issues.
% , where the memory required to hold the weights and intermediate activation data saved for back-propagation grows linearly w.r.t the number of candidates in the search space. 
However, the bottleneck still remains, \emph{i.e.}, the number of candidate paths grows exponentially \emph{w.r.t.} the number of hyper-parameter types, because each combination of different hyper-parameter types (\emph{e.g.}, convolution kernel size, and quantization precision) is considered as an instance of the search candidate. As more hyper-parameters are needed in hardware-architecture co-design, the GPU memory usage can grow orders of magnitude higher, beyond the capacity of many acceleration platforms.

One-Shot Architecture Search (One-Shot \DARTS)~\cite{dong2019one} 
are among the few attempts to alleviate the memory pressure of 
\DARTS~by activating only one candidate path during the SuperNet training. As such, the GPU memory usage can be reduced to the level same as that of training one single neural architecture because only one candidate path is activated in the forward inference and back propagation process. However, the method does not reduce the size of the SuperNet, which still needs to be stored off the GPU memory if too large. In such a scenario, frequent communication between GPU and CPU is needed to load the candidate paths, resulting in significant search time overhead. 
% For example, a one-pass loading of the candidate paths in the SuperNet built for the search space in existing RL-NAS frameworks~\cite{yang2020co} to a RTX-3080 GPU card with 256 Gbps bandwidth would require 0.93 seconds, and to complete the training the SuperNet they need to be loaded enormous times. 

Fundamentally speaking, the major cause for memory usage explosion in \DARTS~is the exponential growth of search space. In vanilla \DARTS, every possible candidate path is added to the search space to form a homogeneous SuperNet. However, not all candidates are reasonable choices. For example, higher bit precision is typically needed by layers closer to the input or output. 
%because different layers require candidates with different properties.
As such, naively including all possible candidates for every layer leads to a significant waste of memory and computation. It is important to prune these unwanted candidates before initiating \DARTS. \textbf{The challenge then is how to find a subspace in the designated search space that (1) fits in the provided acceleration platform with bounded GPU memory, and (2) contains the optimal neural architecture.}

\begin{figure}[h]
%\vskip 0.2in
\begin{center}
\centerline{\includegraphics[trim=0 145 420 0, clip, width=0.8\linewidth] {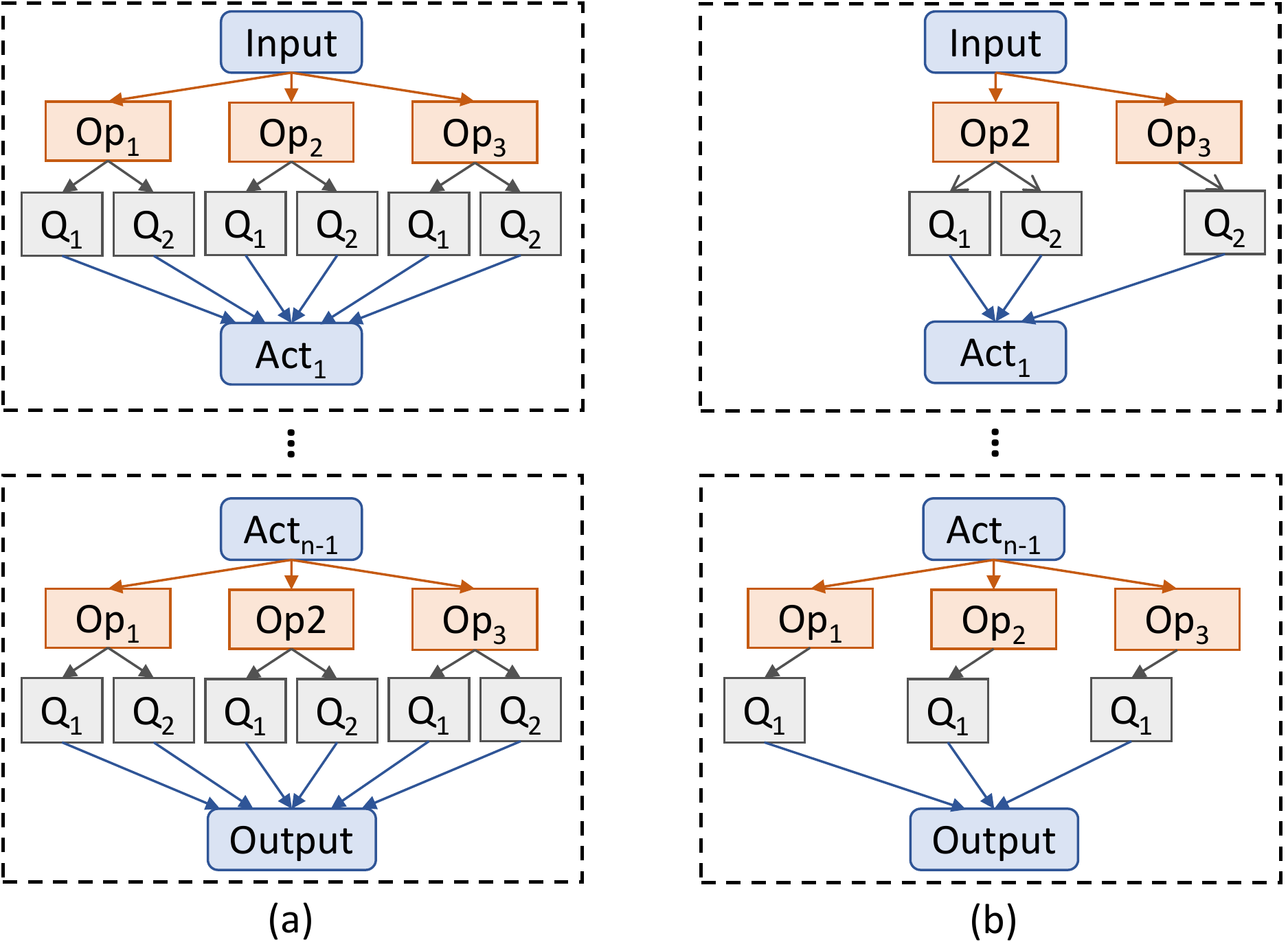}}
\vspace{-0.4cm}
\caption{Conceptual illustration of NAS search space pruning. (a) SuperNet created from the original search space, and (b) reduced SuperNet created by RL-aided pruning.}
% \todo{This figure is generated from testcode/CIM/9874/plot/crossbar.pptx}
\label{figure:idea}
\vspace{-0.8cm}
\end{center}
\end{figure}

In this work, we propose \pName, an effective solution to reduce the GPU memory usage of \DARTS. It prunes the target search space with the guidance of RL, using a method with iterative exploration/exploitation phases. As shown in Fig.~\ref{figure:idea}, the original search space for vanilla \DARTS~includes an identical set of candidate operations across different layers. In each exploration phase, by analyzing high-performance architectures generated by the controller, RL learns to better identify a subset of candidates in each layer that are promising to offer desired performance. Then in the following exploitation phase, a \DARTS~process identifies the best architecture in the RL-identified search subspace. 

We conduct experiments on the CIFAR-10 and ImageNet datasets with large search spaces using a workstation with two RTX-2080Ti GPU cards (12 GB memory). \pName~can achieve strong empirical results on both datasets, achieving up to 3.41\% lower test error and 
39\% lower arithmetic intensity~\cite{li2020edd} with 2.5x speedup compared with a state-of-the-art RL based NAS implementation~\cite{lu2019neural}. On the other hand, 
the \DARTS~\cite{li2020edd} and One-Shot \DARTS~\cite{dong2019one} methods cannot finish due to OOM and excessive search time issues, respectively. 
\todo{The superiority of \pName~over existing \DARTS~methods is further confirmed as the resultant DNN models share similar performance with other state-of-the-art NAS framewors.}
% The superiority of \pName~over existing \DARTS~methods is further confirmed when the search space is scaled down. \pName~achieves comparable accuracy and arithmetic intensity with reduced search time and memory usage compared with the two \DARTS~methods. 
To the best of our knowledge, \pName~is the first \DARTS~framework that can efficiently handle large search spaces with bounded memory usage. 

% The remainder of the paper is organized as follows. We first review background knowledge of RL-NAS, \DARTS, and \DARTS~variants in Section~\ref{sect:background}. We then introduce the \pName~framework in detail and analyze its memory efficiency in Section~\ref{sect:proposed}. Experimental results for both the CIFAR-10 and ImageNet datasets to show the effectiveness of \pName~are provided in Section~\ref{sect:experiment} and concluding remarks are given in Section~\ref{sec:con}.

% \todo{I want to focus on the fact that the memory usage for differentiable NAS grows exponentially when adding different choices for in memory usage. ProxylessNAS relieved it but didn't cut the exponential growth.}

\section{Background}\label{sect:background}
Neural Architecture Search (NAS) aims to explore through a pre-determined set of neural architectures (\emph{i.e.}, search space) and select one neural architecture that can offer optimal performance. In this section, we briefly review two different
categories of NAS methods and their variants, as our method effectively bridges these two methods. 

\textbf{RL-based Neural Architecture Search}\label{sec:RL}
(RL-NAS)~\cite{zoph2016neural, jiang2019accuracy, jiang2020hardware, jiang2020device, yang2020co, yan2021uncertainty} is well known for its memory efficiency in searching neural architectures. It is composed of three key components, a \emph{controller}, a \emph{trainer}, and an \emph{evaluator}. In one iteration (episode) of RL-NAS, the \emph{controller} predicts a neural architecture from the search space; the \emph{trainer} trains the predicted neural architecture, named child network, from scratch on a training dataset for a number of epochs, and the \emph{evaluator} collects the figures of merit (FOM), \emph{e.g.}, latency and/or energy consumption of the child network and its test accuracy of the trained child network on test dataset. The FOM then forms a \emph{reward} function, based on which the \emph{controller} is updated so that it can predict neural architectures with higher FOM.
The neural architecture that offers the highest \emph{reward} among all the searched neural architectures is presented as the search result.

% The termination criteria for this iterative method is that either the \emph{controller} repeatedly predicts the same child network or the number of predicted architectures exceeds a predefined threshold (episode limit). The neural architecture that offers the highest \emph{reward} among all the searched neural architectures is presented as the search result.

% XX Need to mention retrain; epoches; episodes; 

\textbf{Differentiable Neural Architecture Search} (\DARTS) is prevailing for its ability to search neural architectures in a time-efficient manner. One typical way of \DARTS~is to construct and directly train an over-parameterized network, called SuperNet, that contains all possible candidate paths. Each candidate path is associated with an \emph{architecture weight}. 
% This relaxes the NAS task from searching through a discrete search space to an optimization problem in a continuous space. More importantly, 
The \emph{architecture weights} are differentiable with respect to the loss function of the SuperNet, so the \emph{architecture weights} can be optimized with respect to the loss function by gradient descent. For a specific training task, 
% a two-step training approach is adopted, where 
the neural network weights and the \emph{architecture weights} are updated alternately. The final chosen DNN is composed of the candidates with maximum \emph{architecture weights} in each layer. Some variants of \DARTS~\cite{vahdat2020unas} incorporate RL algorithms in the search process to estimate gradients for non-differentiable objectives and help broaden the application field of \DARTS. However, the adoption of RL does not help reduce the memory usage of \DARTS.

% \subsection{\DARTS~Variants}

% Although \DARTS~can reduce the search time to tens of GPU hours, this method suffers from poor scalability, low flexibility, and unstable convergence issues. Variants of \DARTS~have been proposed to address these issues. 

The scalability issue is a major concern when using \DARTS. \DARTS~is not scalable because the introduction of SuperNet requires the weight values and architecture parameters of the whole search space to be stored in memory. One-Shot Architecture Search~\cite{ dong2019one} alleviates this issue by reducing the training process from requiring the whole SuperNet in the forward and backward path to only including one selected candidate path of the SuperNet in one iteration of training. This greatly reduces the requirement of both memory usage and computational power. However, One-Shot Architecture Search does not reduce the size of the SuperNet, which grows exponentially \emph{w.r.t.} the number of hyper-parameters. When it grows beyond the memory capacity of GPUs, excessive communication between GPU and CPU is needed which leads to a much longer search time. 
% Furthermore, the memory needed to store the weights of the SuperNet is not scalable with the growth of the SuperNet, \emph{i.e.}, the design space.

% Hardware-aware differentiable NAS~\cite{wu2019fbnet} and differentiable NAS for hardware-neural architecture co-design~\cite{li2020edd} adopts the SuperNet topology and defines the performance, \emph{e.g.}, energy and latency, of the SuperNet as the weighted sum of the performance of each candidate. The hardware performance and the evaluation loss are then combined into a new loss function to update the \emph{architecture weights}.

% \begin{equation}
%     \begin{split}\label{eq:proxy}
%         NT  & = \prod_{i=0}^{T} D_i \cdot \left( \sum_{l = 0}^{L}  CO_l (\alpha CI_l \cdot K^2)\right) + C
%     \end{split}
% \end{equation}

% ProxylessNAS~\cite{cai2018proxylessnas} reduced DARTS memory usage by only activating one search candidate each time. As shown in Eq.~\ref{eq:proxy}, ProxylessNAS reduced memory usage used to save output activation data to constant. However, the memory usage used for saving the weight data is not reduced, thus still suffering from an exponential memory usage increase.

% ProxylessNAS reduces this to:

% \begin{equation}
%     N_p = \alpha C_i \times C_o \times N \times M \times \sum_{i=0}^{S} K_i^2 + \beta C_o \times (W_o \times H_o)
% \end{equation}
\section{RL Aided Differentiable NAS}\label{sect:proposed}

Below, we first give an overview of \pName, then describe its key components. We finally analyze the memory usage of \DARTS~and show how \pName~can reduce it.

\subsection{Overview}

In this work, we propose \underline{R}einforcement learning \underline{A}ided \underline{D}ifferentiable neural \underline{AR}chitecture \underline{S}earch (\pName). 
Different from existing works that focus on reducing the implementation overhead of \DARTS, we directly prune the search space so that the GPU memory usage can be bounded. The fundamental challenge here is to identify the promising subspace in the entire search space efficiently, towards which no simple heuristics exist. 

The most straightforward approach is to divide the designated search space into a group of non-overlapping subspaces, each of which can fit in the GPU memory. We can then perform \DARTS~in each subspace and find the best architecture~\cite{zhao2021few}. The issue with this method is that it only reduces the number of subspaces needed for search by a constant factor thus the search time is the same order of magnitude as RL-NAS.%but cannot reduce the order of magnitude of the search iteration.}

Inspired by the RL-NAS, we adopt an RL-based scheme to quickly prune undesired candidate paths and predict a promising subspace for \DARTS~so that it no longer needs to fully explore the entire search space. 
 
There are two potential issues for adopting such a pruning scheme, however. First, the RL process is known to be slow. Second, if RL happens to miss a promising candidate, the following \DARTS~will never be able to add it back. To tackle these issues, we devise a search process that interleaves the exploration/exploitation phases, (see in Alg.~\ref{alg:all}). Specifically, each exploration phase (detailed in Section~\ref{sec:support}) predicts a set of promising neural architectures $\mathbf{SN}$. Then, the following exploitation phase (detailed in Section~\ref{sec:build}) builds a SuperNet based on $\mathbf{SN}$, reflecting a pruned search space, and conducts a \DARTS~search. The two phases are carried out iteratively until one of the following stopping criteria is met: 1) the search \emph{reward} is greater than a preset \emph{target} or 2) the number of episodes exceeds a predefined limit. In our experiments, we find that \pName~typically converges in 50 episodes. 
The implementation details are discussed below.

 %For a specific task, \textit{i.e.}, searching for the best architecture given dataset $\mathbf{D}$ and design space $\mathbf{S}$, the \emph{RL supporter} aims to predicting 

%the \emph{RL supporter} firstly predicts $N$ neural architectures $\mathbf{A}_i, i=1,...,N$, from $\mathbf{S}$, each with a designated reward $\mathbf{R}_i$. The \emph{RL supporter} then starts a re-training process to update the reward $\mathbf{R}$. Undesired architectures are then pruned, and the remaining architecture set $\mathbf{F}$ is used by the \emph{SuperNet builder} to build a SuperNet. The \emph{DARTS searcher} then identifies an optimal architecture from the SuperNet. If the identified architecture meets the target, the search stops. Otherwise, another $N$ neural architecture is predicted by RL and added to $\mathbf{A}$. Similar steps are performed iteratively until a neural architecture that meets the target is identified.

% \begin{figure}[h]
% %\vskip 0.2in
% \begin{center}
% \centerline{\includegraphics[trim=0 0 320 0, clip, width=0.8\linewidth] {figures/flow.pdf}}
% \vspace{-0.4cm}
% \caption{Flowchart of \pName~algorithm. The blue blocks indicate major operations, the orange blocks indicate other operations, and the green blocks show resultant output data of an operation.}
% % \todo{This figure is generated from testcode/CIM/9874/plot/crossbar.pptx}
% \label{figure:flow}
% \vspace{-0.8cm}
% \end{center}
% \end{figure}

\begin{algorithm}[h]
\caption{\pName~($S$, $D$, $N$, $P$, $Ep$, $Tar$) $\rightarrow C $}
\begin{algorithmic}[1]\label{alg:all}
\STATE // input: search space $\mathbf{S}$, target task dataset $\mathbf{D}$, $\#$ of episodes in exploration phase $N$, $\#$ of architectures to re-train $P$, maximum searching iteration $Ep$, and the target performance $Tar$;
\STATE // output: best neural architecture identified $C$;
\STATE Initialize empty result set, $\mathbf{R}$; 
\STATE Initialize RL search process using $\mathbf{S}$ and $\mathbf{D}$;
% \STATE Run RL for $N$ episodes;
\WHILE{$reward < Tar$ $\mathbf{and}$ $\#$ of iteration $< Ep$}
    \STATE \cmtColor{//\textit{Exploration phase: }}
    \STATE Continue to run RL search for $N$ episodes. It generates $N$ reward and architecture pairs $\mathbf{R}_i = \{<r_i, a_i> | i = 1, .., N\}$;
    \STATE Append $\mathbf{R}_i$ to $\mathbf{R}$. ($\mathbf{R} \leftarrow \mathbf{R} \cup \mathbf{R}_i$);
    \STATE Identify the top $P$ rewards in $\mathbf{R}$;
    \STATE Re-train the $P$ architectures identified;
    % If any architecture has been retrained previously, skip its retraining;
    \STATE Update $\mathbf{R}$ with the new rewards after re-training;
    \STATE Initialize an empty set of Neural Architecture $\mathbf{SN}$;
    \FOR{$<r, a>$ in $\mathbf{R}$ in descendent order of $r$}
        % \STATE Identify the corresponding architecture $a$ of $r$ in $\mathbf{A}$;
        \STATE Add $a$ to $\mathbf{SN}$ and build a SuperNet from $\mathbf{SN}$;
        \IF{Memory usage $>$ GPURAM capacity}
            \STATE Remove $a$ from $SN$; Break;
        \ENDIF
    \ENDFOR
    \STATE \cmtColor{\textit{// Exploitation phase: }}
    \STATE Build a \DARTS~search scheme based on $\mathbf{SN}$; 
    \STATE Perform \DARTS~and keep track of the best architecture identified so far $C$ and gather the $reward$ of $C$;
    
\ENDWHILE
\end{algorithmic}
\end{algorithm}

\subsection{Exploration Phase}\label{sec:support}
This phase aims to identify promising architectures in the search space so that the following exploitation phase can focus on a pruned space formed by these architectures instead of the entire search space. 

Similar to a conventional RL-NAS as discussed in Section~\ref{sec:RL}, We use a  \emph{controller}, a \emph{trainer}, and an \emph{evaluator} for the exploration phase. 
However, different from the RL-NAS that trains the \emph{controller} until an optimal architecture is found, which can be extremely slow, each time this phase only runs for $N$ episodes (continuing from the previous run) and then predicts a set of architectures $\mathbf{SN}$. The episode number $N$ is specified by the user. Smaller $N$ means higher granularity, thus offering higher performance but requiring longer search time. 
% We find through experiments that setting $N=10$ generally provides the best balance between time efficiency and performance. 
% The number of architectures in $\mathbf{SN}$, is the maximum number of architectures to build a SuperNet (details in Section~\ref{sec:build}) that can be accommodated in the memory. 

Reckoning the fact that the performance of RL in its early stages may not be good, we keep track of all the architectures and the associated rewards identified in the exploration phase. At the end of the $t^{th}$ exploration phase, a total of $N\cdot t$ architectures and rewards are explored and \todo{the top architectures in them} form the pool to identify the most promising subset $\mathbf{SN}$. 

% A naive approach to generate $\mathbf{SN}$ is to simply select the top $|\mathbf{SN}|$ architectures with the highest rewards from the $N\cdot t$ architectures. 
Moreover, the \emph{trainer} only trains the predicted architecture for a limited number of epochs for the sake of time efficiency. Thus, the rewards gathered by the \emph{evaluator} (described in Section~\ref{sect:reward}) are inaccurate, and selecting the top ones based on such inaccurate rewards can easily miss out on good candidates. 
% In RL-NAS, the promising architectures are re-trained until convergence. Yet it is not feasible in our case to retrain the architecture from every episode, which will lead to significant time overhead. 
To address this issue, we take a two-step pruning approach. We first identify $P$ architectures with the highest rewards out of the $N\cdot t$ architectures in the pool as the candidates, where $P$ is specified by the user. These $P$ architectures are then trained to converge to get their exact rewards. Apparently, a larger $P$ provides better performance (with more precise rewards) at the cost of longer search time. $P$ is chosen in early experiments so that the execution time of the exploration and exploitation phase are equal and can thus be executed in parallel without waiting for each other's output. 
% Note that if any of these $P$ architectures have already been retrained previously, \emph{i.e.}, it appears as a candidate in the previous exploration phases, it does not need to be retrained as the reward is already exact. The reward information for the retrained architectures in the pool is updated as well.
With the $P$ candidate architectures and their exact rewards, we can further identify a subset of them with the highest rewards to form $\mathbf{SN}$, which are then used in the exploitation phase to form a SuperNet and identify the best architecture.

%Note that if the pool grows too big to cause memory issues, then instead of adding new architectures to $\mathbf{A}$, they will replace those with the lowest rewards. 
%A new SuperNet is built on the new $\mathbf{F}$ and tested for memory usage. This approach is applied iteratively until the memory constraint is met.

% Finally, $\mathbf{F}$ is sorted in descendent order where the neural architecture with the best reward serves as the first item of $\mathbf{F}$.

\subsection{Exploitation Phase}\label{sec:build}

In the exploitation phase, we first take in the neural architecture set $\mathbf{SN}$ generated in the exploration phase to build a SuperNet that is under GPU memory capacity and reflects the pruned search space. 

\begin{figure}[h]
%\vskip 0.2in
\begin{center}
\centerline{\includegraphics[trim=0 250 430 0, clip, width=0.8\linewidth] {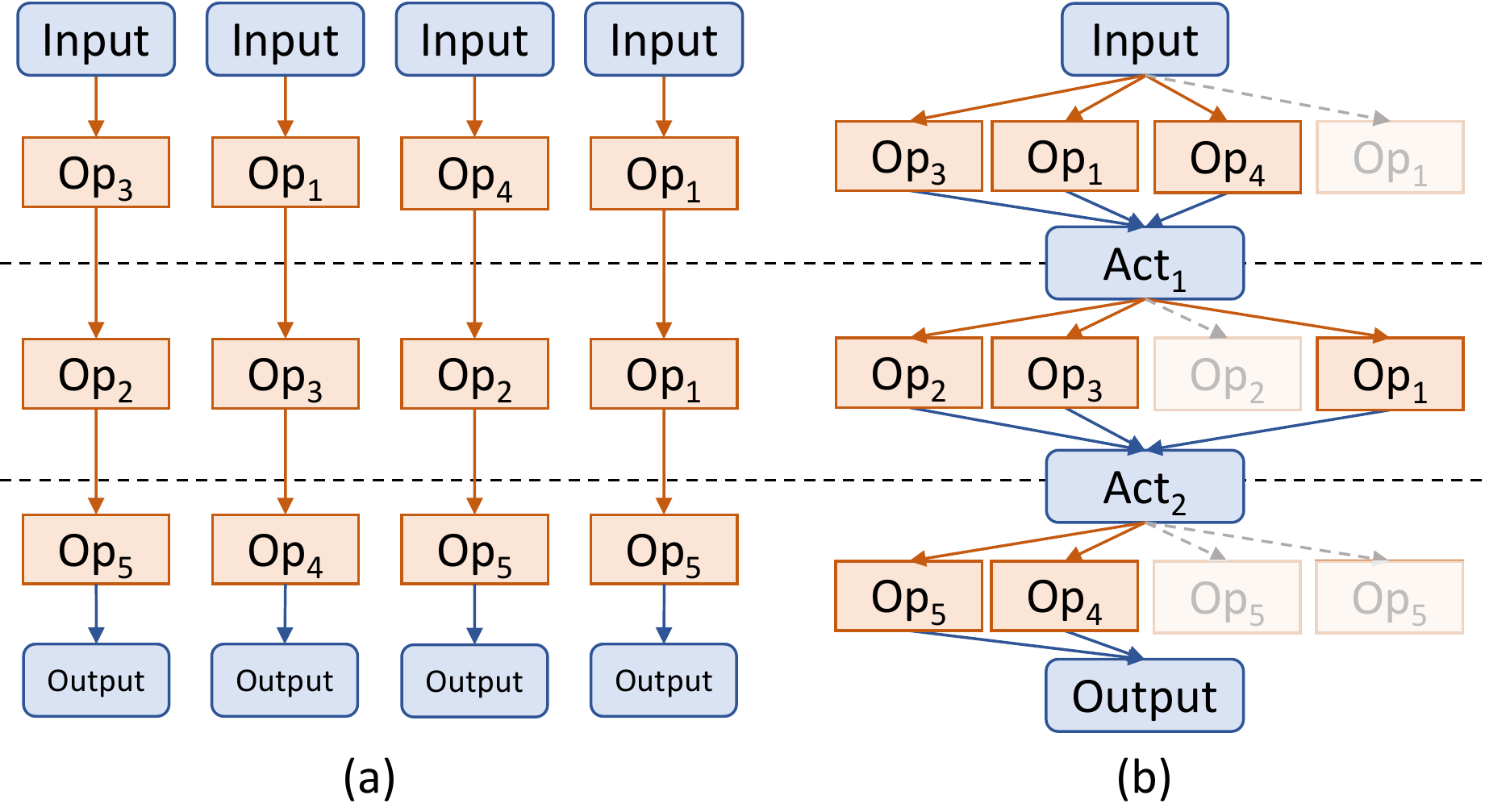}}
\vspace{-0.4cm}
\caption{Illustration for SuperNet building. (a) Architectures generated by the exploration phase, and (b) SuperNet built upon the architectures in (a). In each layer, candidate operations are added to the SuperNet and the redundant ones are discarded. \emph{E.g.}, in layer 3, $Op_4$ and $Op_5$ are added to the SuperNet and the repeated $Op_5$'s are not.}
\label{figure:space}
\vspace{-0.8cm}
\end{center}
\end{figure}

\DARTS~search space is organized in a layer-by-layer manner. In each layer, every candidate operation is added into the SuperNet, associated with an \emph{architecture weight}. Different layers are then stacked on top of each other to form a SuperNet. There are $|\mathbf{SN}|$ candidates in $\mathbf{SN}$ and assume that each candidate has $L$ hyper-parameters.  Then they can form a 2D matrix of $L\times |\mathbf{SN}|$, as shown in Fig.~\ref{figure:space} (a). In each layer, redundant candidates are removed and a compact SuperNet (Fig.~\ref{figure:space} (b)) is formed on which a differential architecture search can be conducted to identify the best architecture.

%If the SuperNet exceeds the memory capacity, 
%then we iteratively remove the architectures in $\mathbf{F}$ with the lowest reward until the resulting SuperNet can fit the memory. 

\subsection{Reward Evaluation}\label{sect:reward}

We use \pName~to search for neural architectures that are both accurate and hardware efficient. 
% Our goal is to demonstrate the effectiveness of \pName, and we do not want to add more platform configuration complications (\emph{i.e.}, simulation tools) to \pName. Thus, 
In this paper, we use the widely adopted arithmetic intensity~\cite{li2020edd} indicator, number of total MAC operations, as our FOM to indicate the hardware performance of a certain neural architecture.  Note that it is widely acknowledged that, although arithmetic intensity does not necessarily represents model latency, it is an appropriate indicator of energy consumption~\cite{lu2019neural}. Accordingly, our framework identifies a DNN architecture that is both accurate and energy-efficient. It is understood that other metrics can also be used if desired.

Moreover, in quantized DNNs, the cost of MAC operations for different bit precisions is not the same in terms of arithmetic intensity. Thus, assuming integer MAC units are used in quantized DNN accelerators, we define adjusted number of operations (AOPS) as an arithmetic intensity indicator for quantized DNNs. AOPS can be defined as:
% In integer multiplications, the arithmetic intensity of one operation is related to the bit precision of both operands, and AOPS can be defined as:
\vspace{-0.1cm}
\begin{equation}\label{eq:op}
    \vspace{-0.1cm}
    AOPS = MACOPS \times Activation\ bits \times Weight\ bits
    % \vspace{-0.2cm}
\end{equation}
where the number of MAC operations, bit precision of weights, and input activations are taken into consideration.

The performance evaluation metric (reward function) used in the search process can be defined as:
\vspace{-0.1cm}
\begin{equation}\label{eq:reward}
\vspace{-0.1cm}
    Reward = \alpha\times Acc + (1-\alpha) \times (1 - (AOPS - \beta) / \gamma)
\end{equation}
where $\alpha$, $\beta$, and $\gamma$ are user-defined normalization parameters.

\subsection{Memory Usage Analysis}
\label{sec:complexity}

% \begin{table}[h]
%     \caption{Symbol representations used in this section.}
%     \label{tab:symbols}
%     \vspace{-0.2cm}
%     \begin{center}
%     \begin{tabular}{cl}
%     \toprule
%     Symbol & Meaning \\
%     \midrule
%     $B$     & batch size   \\                 
%     $CI$    & \# of input channels \\          
%     $CI$ & \# of input channels\\
%     $D_i$ & \# of choices for a hyperparameter\\
%     $WI$    & input feature map width      \\
%     $HI$    & input feature map height     \\
%     $WO$    & output feature map width     \\
%     $HO$    & output feature map height    \\
%     $K$     & kernel size                  \\
%     $T$   & \# of hyper-parameter types \\
%     $NW$  & \# of weights\\
%     $\eta$& unit memory usage for weight \\
%     $NA$  & \# of activations\\
%     $\theta$ & unit memory usage for activation\\
%     $NM$  & normalized memory usage\\
%     $L$   & \# of layers\\
    
%     \bottomrule
    
%     \end{tabular}
%     \end{center}
% \end{table}
\DARTS~is memory-hungry because its memory usage increases drastically with the expansion of the search space. Below, we show that the memory requirement of \DARTS~grows exponentially \emph{w.r.t.} the number of hyper-parameters. We also show that \pName~can change this growth to constant.

Without loss of generality, we analyze a \DARTS~search space with $L$ convolutional layers and $T$ different types of hyper-parameters (\textit{e.g.}, kernel size, bits of quantization) for each layer. For the $i^{th}$ hyper-parameter type, there are $D_i$ candidates to choose from. 
% To simplify the representations, each layer in this reasoning is considered a quantized convolution layer and hyper-parameters are hardware-related specifications, \textit{e.g.}, number of bits for weights and activations.

In the $l^{th}$ layer of the SuperNet, it gets an input sized $B \times CI_l \times WI_l \times HI_l$, where $B$ is batch size, $CI$ is the number of input channels, $WI$ and $HI$ are input width and height. It also creates an output sized $B \times CO_l \times WO_l \times HO_l$.
% , where the representations are similar to the input
This layer uses a kernel size of $K$.

In the $l^{th}$ layer, the total number of convolution weights and gradient data can be represented by $NW_l$ and the total number of output activation data can be represented by $NA_l$, respectively.
\vspace{-0.2cm}
\begin{align}
    NW_l =  CI_l \times CO_l \times K^2 \times \prod_{i=0}^{T} D_i \\
    NA_l =  B \times CO_l \times WO_l \times HO_l \times \prod_{i=0}^{T} D_i
\end{align}

Finally, the total memory usage of the whole SuperNet can be calculated by:
\vspace{-0.2cm}
\begin{equation}
    \label{eq:allMem}
    \begin{split}
        & NM = \sum_{l = 0}^{L} \eta \times NW_l + \theta \times NA_l\\ 
        & = \prod_{i=0}^{T} D_i \cdot \left( \sum_{l = 0}^{L}  CO_l (\eta CI_l \cdot K^2 + \theta B \cdot WO_l \cdot HO_l)\right) \\
    \end{split}
\end{equation}
where $\eta$ and $\theta$ are coefficients related to the data representations of weights and activations. 
% \begin{equation}
%     \begin{split}\label{eq:allMem}
%         NM   = \prod_{i=0}^{T} D_i \cdot \left( \sum_{l = 0}^{L}  CO_l (\eta CI_l \cdot K^2 + \theta B \cdot WO_l \cdot HO_l)\right)
%     \end{split}
% \end{equation}

It can be observed from Eq.~\ref{eq:allMem} that memory usage grows exponentially \emph{w.r.t.} the number of hyper-parameter types ($T$), which todo{blocks} the exploration of more different types of hyper-parameters. For example, with the search space specified in Table~\ref{tab:setupCIFAR}, the memory usage grows up to 112 GB. This is beyond the capacity of many acceleration platforms. 

The proposed \pName~framework efficiently reduces the memory usage \emph{w.r.t.} hyper-parameter types from exponential growth to a user-defined constant. \todo{As shown in Alg.~\ref{alg:all}, neural architecture set $\mathbf{SN}$ used to build the SuperNet is a subset of $P$ candidates, so $|\mathbf{SN}| \leq P$. Let the memory usage of a single path be $M_{sp}$. The memory usage of this SuperNet is $|\mathbf{SN}| \times M_{sp} \leq P \times M_{sp}$.}
% (see Eq.~\ref{eq:pMem})
Thus, the memory usage of \pName~is bounded by a constant independent of the growth of the design space.
% The benefit of this reduction becomes more apparent as the number of hyper-parameter types grows larger. On the other hand, as only the inferior search space is pruned, the quality of the architectures identified by RADARS will not degrade, as will be demonstrated in the experiments.{}

% \begin{equation}
%     \begin{split}\label{eq:pMem}
%         PT  & \leq Y \cdot \left( \sum_{l = 0}^{L}  CO_l (\alpha CI_l \cdot K^2 + \beta B \cdot WO_l \cdot HO_l)\right)
%     \end{split}
% \end{equation}

\section{Experiments}\label{sect:experiment}
We apply our proposed \pName~to two different tasks: CIFAR-10~\cite{krizhevsky2009learning} and ImageNet~\cite{deng2009imagenet} classification. All search experiments discussed in this paper are performed on a platform with two RTX-2080Ti GPU cards whose memory capacity is \textbf{12 GB}. 
We should point out that the platform used in our experiment is a lower-end one and is used to illustrate the potential advantage of the proposed method. Some of the 
algorithms that fail to run on our platform may be able to run on higher-end platforms because they have access to larger GPU memory. However, being able to reduce the memory requirements is always needed because of the ever-increasing size of neural networks~\cite{li2020edd} and the accompanied exponential growth of search space. 

We compare our method with three state-of-the-art NAS approaches: one state-of-the-art RL-NAS approach, QuantNAS~\cite{lu2019neural}, and two \DARTS~methods, \DARTS~\cite{li2020edd} and One-Shot \DARTS~\cite{dong2019one}, using their respective implementations but on larger search spaces. If any of the methods need more than the GPU capacity, an out-of-memory (OOM) error is reported. If any of the methods cannot converge in $10\times$ of the search time of the fastest method among them, its search process is terminated and an out-of-time (OOT) error is reported. For all data presented in this paper, the \textit{accuracy} is an averaged result out of three different runs, and \textit{time consumption} refers to the time needed to identify the optimal architecture, excluding the time needed to train the identified architecture from scratch. For user-defined hyper-parameters in Alg.~\ref{alg:all}, we set $N = 10$ and $P = 6$. Other parameter settings are provided below where appropriate.

\subsection{Experiments on CIFAR-10}

We demonstrate the effectiveness of \pName~by searching for an optimal quantized CNN for CIFAR-10. The fixed design parameters and hyper-parameters included in the search space are shown in Table~\ref{tab:setupCIFAR}.
%Thus, \pName~is needed to perform efficient search. 

\begin{table}[t]
    \caption{Quantized CNN for CIFAR-10 search setups. Upper half: configurations fixed to be the same among all architectures; lower half: hyper-parameters to be searched.}
    \vspace{-0.4cm}
    \label{tab:setupCIFAR}
    \begin{center}
    \begin{tabular}{ll}
    \toprule
    {Hyper-Parameter type} & Settings \\
    \toprule
    Block type                      & Quantized Convolution \\
    \# of convolution layers        & 6               \\
    \# of channels per layer   & [64, 64, 128, 128, 256, 256] \\
    Stride                          & [1, 2, 1, 2, 1, 2] \\
    % \# of fully connected layers    & 2              \\
    % Hidden neuron size              & 512            \\
    \midrule
     Kernel size choices             & (1, 3, 5, 7)   \\
    \# of integer bits choices       & (1, 3)        \\
    \# of fraction bits choices      & (1, 3, 6)     \\
    \bottomrule

    \end{tabular}
    \end{center}
    \vspace{-0.8cm}
\end{table}

For reward specifications, we use $\alpha=0.5$, $\beta=0$ and $\gamma=10^{9}$ so that both accuracy and $AOPS$ are normalized to $[0,1]$. Different runs of the search are conducted with different specifications and a variety of neural architectures are identified by \pName~and quantNAS.
The results of \pName, as well as those of quantNAS~\cite{lu2019neural}, \DARTS~\cite{li2020edd} and One-Shot \DARTS~\cite{dong2019one} are reported in Table~\ref{tab:CIFAR}. Because of the memory usage explosion, \DARTS~requires $10.7\times$ more memory than \pName, far exceeding the memory capacity and leading to an OOM error. On the other hand, One-Shot \DARTS~requires frequent data movement between the GPU memory and CPU and is still running after 10 days, more than $26\times$ that of RADARS (OOT error). While quantNAS also converges, \pName~achieves 3.41\% lower test error and 39\% lower arithmetic intensity, with $2.5\times$ search time reduction.

\begin{table}[t]
    \caption{Comparison between different methods on CIFAR-10. \pName~achieves higher top-1 accuracy than QuantNAS in less time. \DARTS~and One-Shot \DARTS~cannot generate results: \DARTS~faces out-of-memory (OOM) issue and One-Shot \DARTS~suffers from extremely long search time (OOT).}
    \vspace{-0.4cm}
    \label{tab:CIFAR}
    \begin{center}
    \begin{tabular}{ccccc}
    \toprule
    \multirow{2}{*}{Model}  & Top-1 & AOPS & Time & Memory \\
    & (\%) &  (G) &  (h) & (GB)\\
    \midrule
    QuantNAS~\cite{lu2019neural}    & 84.92     & 4.09 & 22.9 & 1.430\\
    One-Shot~\cite{dong2019one}    & OOT       & OOT  & $>$240  & 1.537\\
    \DARTS~\cite{li2020edd}          & OOM       & OOM  & OOM  & 112.0\\
    \midrule
    \pName~(ours)                   & 88.33     & 2.50 & 9.15 & 10.43\\
    \bottomrule

    \end{tabular}
    \end{center}
    \vspace{-0.6cm}
\end{table}

\begin{figure}[t]
\begin{center}
\centerline{\includegraphics[trim=0 210 450 0, clip, width=0.8\linewidth] {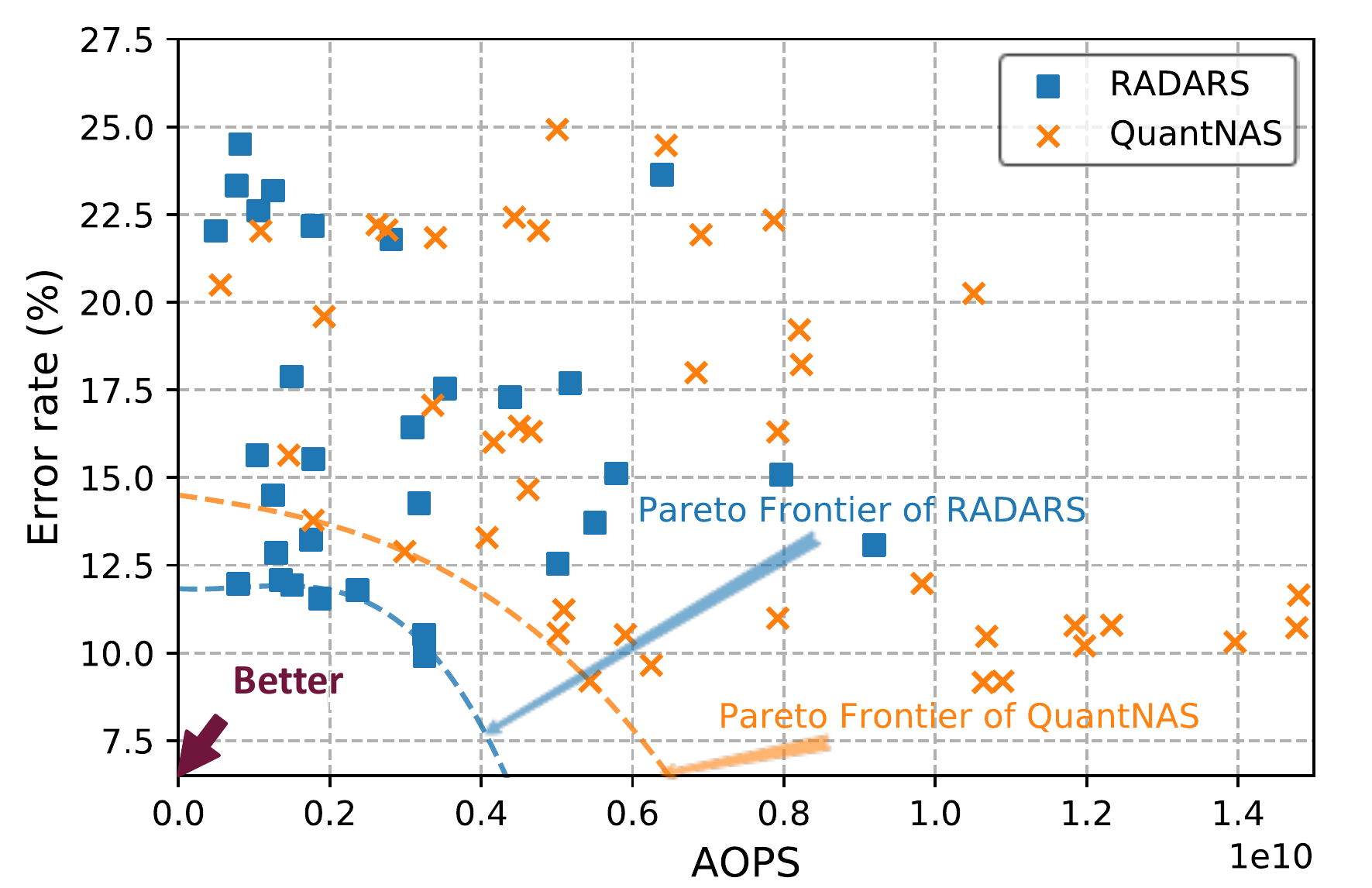}}
\vspace{-0.4cm}
\caption{Comparison between search results of \pName~and quantNAS for CIFAR-10. Each point represents one searched architecture. \pName~achieves superior Pareto frontier.}
\label{figure:pf}
\vspace{-0.8cm}
\end{center}
\end{figure}

Fig.~\ref{figure:pf} shows the FOM of the resulting architectures in terms of accuracy and AOPS by the two methods that are able to complete (\pName~and QuantNAS). In this figure, the x-axis and y-axis represent the AOPS and error rate, respectively. Each rectangle stands for an architecture identified by \pName~and each cross stands for one identified by quantNAS. A solution is better as it moves towards the bottom-left corner. From the results, we can see that by adopting a more efficient search scheme, \pName~can significantly push forward the Pareto frontier between accuracy and AOPS.

\begin{table}[h]
    \caption{MobileNet like search space for ImageNet classification experiment setups. Upper half: configurations fixed to be the same among all searched architectures; lower half: hyper-parameters to be searched.}
    \vspace{-0.4cm}
    \label{tab:setupIN}
    \begin{center}
    \begin{tabular}{ll}
    \toprule
    {Hyper-Parameter type} & Settings \\
    \toprule
    Block type                      & Mobile Invtd Res Block \\
    \# of blocks                    & 6               \\
    \# of channels per block        & [24, 40, 80, 96, 192, 320] \\
    \# of cells per block           & [4, 4, 4, 4, 4, 1]\\
    Stride                          & [2, 2, 2, 1, 2, 1] \\
    % \# of fully connected layers    & 2              \\
    % Hidden neuron size              & 1280            \\
    \midrule
     Kernel size choices            & (3, 5, 7)   \\
    \# of convolution groups           & (3, 6)        \\
    \bottomrule

    \end{tabular}
    \end{center}
    \vspace{-0.8cm}
\end{table}

\subsection{Experiments on ImageNet}

For ImageNet experiments, we focus on learning efficient CNN architectures that can offer high accuracy and at the same time require low arithmetic intensity. Thus, we use MobileNetV2~\cite{sandler2018mobilenetv2} as the backbone. We build a search space similar to the one used in~\cite{cai2018proxylessnas} and use similar search setups as detailed in Table~\ref{tab:setupIN}.

To reduce the search time, similar to existing works~\cite{liu2018darts,liu2018progressive,li2020edd}, we use a proxy dataset, which is a subset of ImageNet with only 100 classes, in the exploration phase of \pName, and use the full ImageNet dataset in the exploitation phase. For the reward function, we use $\alpha=0.9$, $\beta=3\times 10^{8}$ and $\gamma=3\times 10^{8}$. 

The experimental results are shown in Table~\ref{tab:imageNet}. Similar to CIFAR-10 experiments, \DARTS~and One-Shot \DARTS~face OOM and OOT issues respectively, and the proposed \pName~achieves 1.4\% higher top-1 accuracy than QuantNAS using only 54\% of the search time.

\begin{table}[t]
    \caption{Comparison between different methods on ImageNet. \pName~achieves higher accuracy than QuantNAS in less time. One-Shot \DARTS~suffers from extremely long search time (OOT) and \DARTS~aces out-of-memory (OOM).}
    \vspace{-0.8cm}
    \label{tab:imageNet}
    \begin{center}
    \begin{tabular}{cccccc}
    \toprule
    \multirow{2}{*}{Model} & Top-1 & Top-5 & AOPS  & Time  & Mem \\
          & (\%)  & (\%)  & (G)   & (h)   & (GB)   \\
    \midrule
    QuantNAS~\cite{lu2019neural}    & 72.4  & 90.4  & 0.409 & 138   & 6.73\\
    One-Shot~\cite{dong2019one}     & OOT  & OOT     & OOT    & $>$740   & 7.01\\
    \DARTS~\cite{li2020edd}          & OOM   & OOM  & OOM   & OOM   & 58.2\\
    \midrule
    \pName                   & 73.8   & 91.5 & 0.386 & 74    & 11.0\\
    \bottomrule
    
    \end{tabular}
    \end{center}
    \vspace{-0.8cm}
\end{table}

To further study the quality of \pName~produced networks, we also compare the model identified by \pName~with hand-crafted neural network baselines 
% (ResNet-34~\cite{he2016deep} , ShuffleNet~\cite{ma2018shufflenet}, MobileNet V2~\cite{sandler2018mobilenetv2}, and CondenseNet~\cite{huang2018condensenet}) 
and models identified by existing NAS frameworks that require much longer search time 
% (NASNet-A\cite{zoph2018learning} and PNASNET-5~\cite{liu2018progressive} ) 
or larger GPU memory
% (Proxyless-GPU\cite{cai2018proxylessnas})
. As shown in Table~\ref{tab:OtherimageNet}, the model identified by \pName~requires 8x lower arithmetic intensity than ResNet-34 and ShuffleNet, achieves similar accuracy with 30\% lower arithmetic intensity compared with CondenseNet, and achieves 1.8\% higher accuracy with only 30\% higher arithmetic intensity compared with MobileNet V2. Compared with the models identified by existing NAS methods, \pName~achieves lower but comparable accuracy with no less than 20\% arithmetic intensity reduction (\emph{e.g.}, 0.2\% lower accuracy but 30\% lower arithmetic intensity than NASNet-A). This is because, in the \pName~setup, we focus more on energy efficiency in the choice of our search space and thus it is different from 
those used in 
NASNet-A, PNASNET-5, and Proxyless-GPU. On the other hand, though the search spaces of all four NAS frameworks are of similar dimension, \pName~is 650x faster than NASNet-A and PNASNET-5, and uses 30\% lower memory compared with Proxyless-GPU, based on the statistics reported in the respective literature. 

% \todo{We do not offer this result in CIFAR-10 experiments because although there are plenty of work on NAS for quantized neural networks, the experimental settings, search spaces and design objectives (\emph{e.g.} energy consumption and latency) are very different among different works and is thus not feasible to offer a concrete comparison.}

\begin{table}[t]
    \caption{Comparison between the search results of the proposed and baseline methods on ImageNet. Note that the results from existing NAS frameworks are directly obtained from the respective literature ~\cite{zoph2018learning,cai2018proxylessnas,liu2018progressive}, each of which uses a different search space of similar dimension.}
    \vspace{-0.4cm}
    \label{tab:OtherimageNet}
    \begin{center}
    \begin{tabular}{ccccc}
    \toprule
    \multirow{2}{*}{Type} & \multirow{2}{*}{Model} & Top-1 & Top-5 & AOPS \\
    & & (\%)  & (\%)  & (G)\\
    \midrule
    \multirow{2}{*}{Hand}
    & ResNet-34~\cite{he2016deep}                 & 73.3 & 91.4 & 3.600\\
    % & ShuffleNet V2~\cite{ma2018shufflenet}       & 77.8 & 93.3 & 2.300 \\
    % \multirow{2}{*}{crafted}
    & CondenseNet~\cite{huang2018condensenet}     & 73.8 & 91.7 & 0.529 \\
    \multirow{1}{*}{crafted}
    & MobileNet V2~\cite{sandler2018mobilenetv2}  & 72.0 & 91.0 & 0.300 \\
    \midrule
    \multirow{2}{*}{NAS} 
    & NASNet-A~\cite{zoph2018learning}            & 74.0 & 91.6 & 0.564\\
    & PNASNET-5~\cite{liu2018progressive}         & 74.2 & 91.9 & 0.588\\
    identified & Proxyless-GPU\cite{cai2018proxylessnas}     & 75.1 & 92.5 & 0.465\\ %0.465260160\\
    \midrule
    Proposed & \pName~(ours)                      & 73.8 & 91.5 & 0.386 \\%0.386043680\\ %my5
                                %   0.395743328\\
                                %   0.359842400\\
    \bottomrule

    \end{tabular}
    \end{center}
    \vspace{-0.8cm}
\end{table}
% \todo{write 12 GB is text}
% \todo{When discussing tables, you write quote some numbers.}

% \todo{change even in advanced data centers $-->$ average platforms}

To summarize, for the ImageNet dataset, \pName~can efficiently search large hardware-related design spaces, while \DARTS~and One-Shot \DARTS~do not work. On the other hand, \pName~also achieves performance comparable with other NAS baselines with design spaces that are different but of similar dimensions, while using significantly shorter search time and lower memory usage.
\section{Conclusions}
\label{sec:con}
In this work, we propose \pName, an RL-aided \DARTS~framework that can efficiently explore a large hardware-aware neural network search space while keeping the memory consumption scalable. \pName~iteratively uses an RL-based exploration phase to identify the sub-spaces in which the most promising architectures reside, followed by an exploitation phase to search the sub-space. Experiments on CIFAR-10 and ImageNet demonstrate the superiority of \pName~over the state-of-the-art.

%% the bibliography file.
\bibliographystyle{IEEEtran}
{\footnotesize

\bibliography{IEEEabrv.bib, M7_References.bib}
}

\end{document}